\documentclass[10pt,twocolumn,letterpaper]{article}

\usepackage{cvpr}              
\usepackage{algorithm}
\usepackage{algorithmic}
\usepackage[dvipsnames]{xcolor}

\definecolor{cvprblue}{rgb}{0.21,0.49,0.74}
\usepackage[pagebackref,breaklinks,colorlinks,citecolor=cvprblue]{hyperref}

\def\paperID{15783} 
\def\confName{CVPR}
\def\confYear{2024}

\title{A Simple and Effective Point-based Network \\ for Event Camera 6-DOFs Pose Relocalization}

\author{Hongwei Ren\thanks{equal contribution. \dag corresponding author.} , Jiadong Zhu\textsuperscript{*}, Yue Zhou, Haotian Fu, Yulong Huang, Bojun Cheng \dag\\
The Hong Kong University of Science and Technology(Guangzhou)\\
{\tt\small \{hren066,jzhu484,yzhou833,hfu373,yhuang496\}@connect.hkust-gz.edu.cn, bocheng@hkust-gz.edu.cn}
}

\begin{document}
\maketitle
\begin{abstract}
Event cameras exhibit remarkable attributes such as high dynamic range, asynchronicity, and low latency, making them highly suitable for vision tasks that involve high-speed motion in challenging lighting conditions. These cameras implicitly capture movement and depth information in events, making them appealing sensors for Camera Pose Relocalization (CPR) tasks. Nevertheless, existing CPR networks based on events neglect the pivotal fine-grained temporal information in events, resulting in unsatisfactory performance. Moreover, the energy-efficient features are further compromised by the use of excessively complex models, hindering efficient deployment on edge devices. In this paper, we introduce PEPNet, a simple and effective point-based network designed to regress six degrees of freedom (6-DOFs) event camera poses. We rethink the relationship between the event camera and CPR tasks, leveraging the raw Point Cloud directly as network input to harness the high-temporal resolution and inherent sparsity of events. PEPNet is adept at abstracting the spatial and implicit temporal features through hierarchical structure and explicit temporal features by Attentive Bi-directional Long Short-Term Memory (A-Bi-LSTM). 
By employing a carefully crafted lightweight design, PEPNet delivers state-of-the-art (SOTA) performance on both indoor and outdoor datasets with meager computational resources. 
Specifically, PEPNet attains a significant 38\% and 33\% performance improvement on the random split IJRR and M3ED datasets, respectively.
Moreover, the lightweight design version PEPNet$_{tiny}$ accomplishes results comparable to the SOTA while employing a mere 0.5\% of the parameters. 
\end{abstract}    
\section{Introduction}
\label{sec:intro}
Event camera is a type of bio-inspired vision sensor that responds to local changes in illumination exceeding a predefined threshold \citep{lichtsteiner2008128}. Differing from conventional frame-based cameras, event cameras independently and asynchronously produce pixel-level events. Notably, event cameras boast an exceptional triad: high dynamic range, low latency, and ultra-high temporal resolution. This unique combination empowers superior performance under challenging light conditions, adeptly capturing the swift scene and rapid motion changes in near-microsecond precision \citep{posch2010qvga}. Additionally, event cameras boast remarkably low power consumption positioning them as a popular choice for many power-constrained devices. 
Camera Pose Relocalization (CPR) is an emerging application in power-constrained devices and has gained significant attention. It aims to train several scene-specific neural networks to accurately relocalize the camera pose within the original scene used for training.
It is extensively employed in numerous applications, including Virtual Reality (VR), Augmented Reality (AR), and robotics \citep{shavit2019introduction}, all of which are deployed on battery-powered devices and are power-constrained. 
\begin{figure}
\centering
\includegraphics[width=0.65\linewidth]{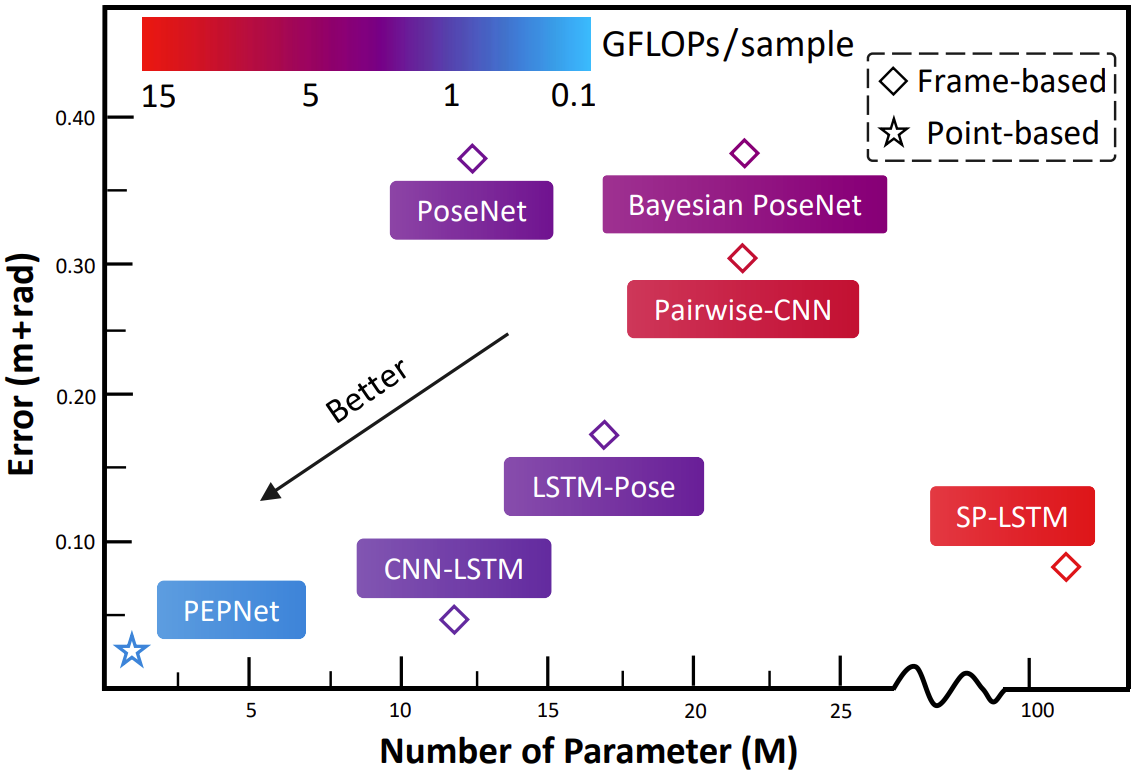}
\caption{The average results using the random split method benchmarked on the CPR dataset \citep{mueggler2017event}. The vertical axis represents the combined rotational and translational errors (m+rad). PEPNet is the first point-based CPR network for event cameras.}
\label{fig: gain of pepnet}
\end{figure}

CPR tasks using event cameras significantly diverge from their conventional CPR counterpart that employs frame-based cameras, primarily due to the inherent dissimilarity in data output mechanisms between these two camera types. Furthermore, events inherently encompass information regarding object motion and depth changes across precise temporal and spatial dimensions attributes of paramount significance within the domain of CPR tasks \citep{rebecq2018esim,gallego2017event}. Regrettably, existing event-based CPR networks often derive from the conventional camera network paradigms and inadequately address the unique attributes of event data. More specifically, events are transformed into various representations such as event images \citep{nguyen2019real}, time surfaces \citep{lin20226}, and other representations\citep{lin20226}, leading to the loss of their fine-grained temporal information. Furthermore, most event-based methods tend to overlook the computational load of the network, only prioritizing elevated accuracy, which contradicts the fundamental design principles of event cameras \citep{gallego2020event}. 

A suitable and faithful data representation is crucial for event cloud processing. Point Cloud is a collection of 3D points $(x,y,z)$ that represents the shape and surface of an object or environment commonly used in lidar and depth cameras \citep{guo2020deep}. The distance $(z)$ is of great meaning to the tasks. 
As for event camera, by treating each event's temporal information as the third dimension, event inputs $(x,y,t)$ can be regarded as points and aggregated into a pseudo-Point Cloud \citep{wang2019space,qi2017pointnet,qi2017pointnet++,sekikawa2019eventnet,ren2023ttpoint,ren2023spikepoint}. However, given that the $t$ dimension of Event Cloud is not strictly equivalent to the spatial dimensions $(x,y,z)$, direct transplantation of the Point Cloud network has not yet exhibited a satisfactory performance advantage in processing event data \cite{wang2019space,ren2023ttpoint}.

\begin{figure}
\centering
\includegraphics[width=0.65\linewidth]{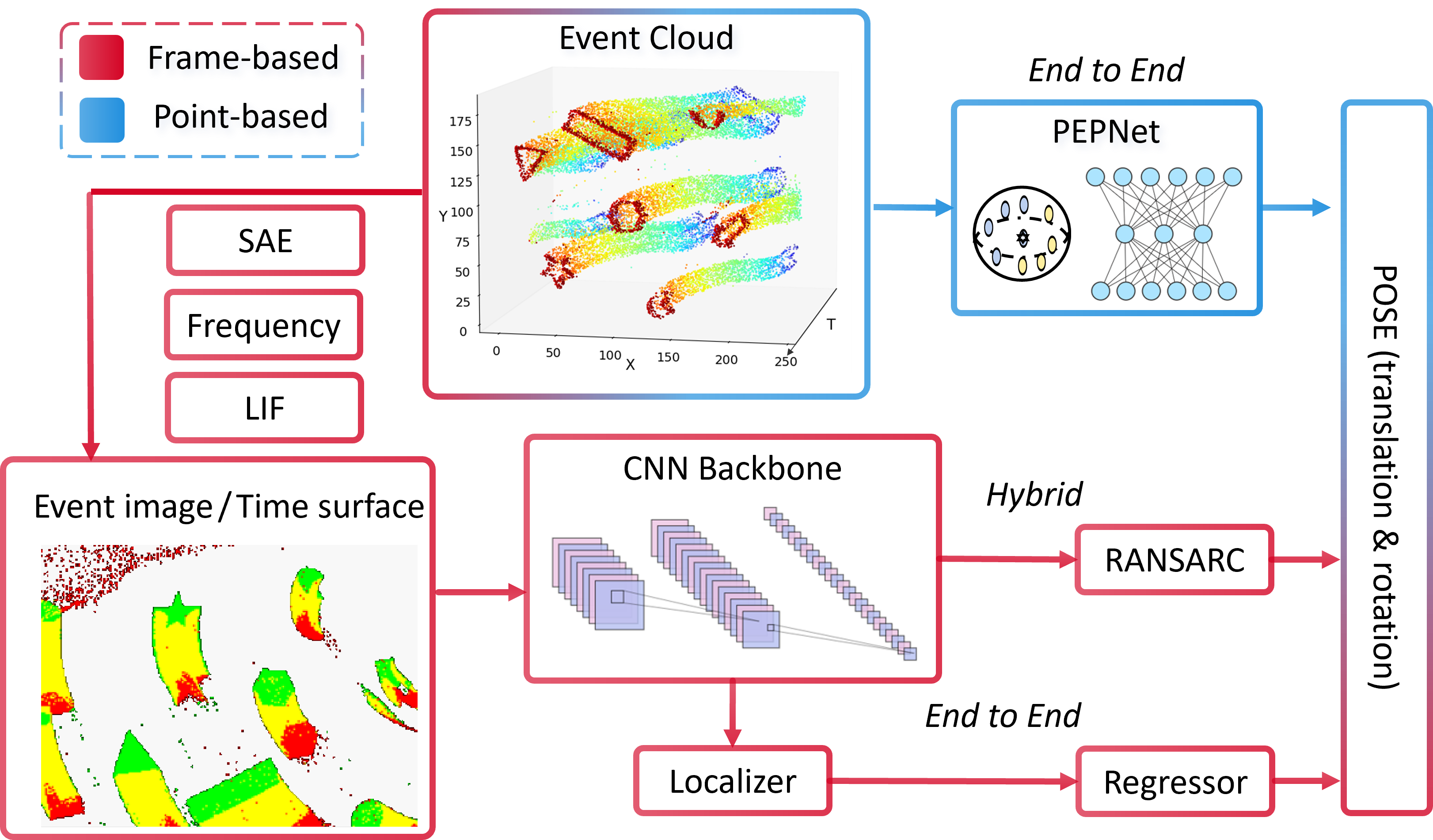}
\caption{Two different event-based processing methods, frame-based and point-based. 
}
\label{figure: two method}
\end{figure}

In this study, we introduce PEPNet, the first point-based end-to-end CPR network designed to harness the attributes of event cameras. A comparison of our performance and method to other frame-based methods is illustrated \cref{fig: gain of pepnet} and \cref{figure: two method}, respectively. Moreover, diverging from other point-based approaches in event data processing \citep{wang2019space, ren2023ttpoint}, PEPNet demonstrates careful attention to detail by systematically assessing the difference between Event Cloud and Point Cloud in its design approach. This approach enables a more precise extraction of spatio-temporal features and facilitates solutions for a spectrum of event-based tasks.
Our main contributions are as follows: First, in the preprocessing stage, PEPNet directly processes the raw data obtained from the event cameras, meticulously preserving the fine-grained temporal coordinate and the order inherent in the event data. Second, PEPNet proficiently captures spatial features and \textbf{implicit temporal} features through its hierarchical structure with temporal aggregation. Subsequently, the \textbf{explicit temporal} feature is processed by the A-Bi-LSTM, thanks to the preservation of the input sequence in previous stages.
As such, this architecture is tailored to accommodate the high temporal resolution and sparse characteristics inherent in event cameras. Thirdly, by restricting ourselves to minimal hardware resources and deliberately avoiding heavy computational modules, 
PEPNet not only attains SOTA results on IJRR \citep{mueggler2017event} and M3ED \citep{chaney2023m3ed} dataset 
but also features a lightweight design that can be executed in real-time.  We hope such an approach could potentially democratize computer vision technology by making it accessible to a wider range of devices and applications in the community of edge computing.


\section{Related Work}
\label{sec:Related Work}
\subsection{Frame-based CPR Learning Methods}
Deep learning, crucial for vision tasks like classification and object detection \citep{lecun2015deep}, has seen advancements such as PoseNet's innovative transfer learning \citep{kendall2015posenet}. Utilizing VGG, ResNet \citep{simonyan2014very,he2016deep}, LSTM, and customized loss functions \citep{walch2017image,wu2017delving,naseer2017deep}, researchers enhanced this approach. Auxiliary Learning methods further improved performance \citep{valada2018deep,radwan2018vlocnet++,lin2019deep}, although overfitting remains a challenge. Hybrid pose-based methods, combining learning with traditional pipelines \citep{laskar2017camera,balntas2018relocnet}, offer promise. DSAC series, for instance, achieve high pose estimation accuracy \citep{brachmann2021visual,brachmann2017dsac}, but come with increased computational costs and latency, especially for edge devices.

\subsection{Event-based CPR Learning Methods}
Event-based CPR methods often derive from the frame-based CPR network. SP-LSTM \citep{nguyen2019real} employed the stacked spatial LSTM networks to process event images, facilitating a real-time pose estimator. To address the inherent noise in event images, \citep{jin20216} proposed a network structure combining denoise networks, convolutional neural networks, and LSTM, achieving good performance under complex working conditions. In contrast to the aforementioned methods, a novel representation named Reversed Window Entropy Image (RWEI) \citep{lin20226} is introduced, which is based on the widely used event surface \citep{mitrokhin2020learning} and serves as the input to an attention-based DSAC* pipeline \citep{brachmann2021visual} to achieve SOTA results. However, the computationally demanding architecture involving representation transformation and hybrid pipeline poses challenges for real-time execution. Additionally, all existing methods ignore the fine-grained temporal feature of the event cameras, and accumulate events into frames for processing, resulting in unsatisfactory performance. 

\subsection{Point Cloud Network}
Point-based methodologies have transformed the direct processing of Point Cloud, with PointNet \citep{qi2017pointnet} as a standout example. Taking a step beyond, PointNet++ \citep{qi2017pointnet++} introduced a Set Abstraction module. While it initially employed a straightforward MLP in the feature extractor, recent advancements have seen the development of more sophisticated feature extractors to enhance Point Cloud processing \citep{wu2019pointconv,zhao2021point,ma2021rethinking,dosovitskiy2020image}. When extending these techniques to Event Cloud, Wang et al. \citep{wang2019space} addressed the temporal information processing challenge while maintaining representation in both the x and y axes, enabling gesture recognition using PointNet++. Further enhancements came with PAT \citep{yang2019modeling}, which incorporated self-attention and Gumbel subset sampling, leading to improved performance in recognition tasks. 
However, existing point-based models still fall short in performance compared to frame-based methods. This phenomenon can be attributed to the distinctively different characteristics of Point Cloud and Event Cloud. 
Event Cloud contradicts the permutation and transformation invariance present in Point Cloud due to its temporal nature. Additionally, the Point Cloud network is not equipped to extract explicit temporal features.
\section{PEPNet}
\begin{figure*}[t]
\centerline{\includegraphics[width=17cm]{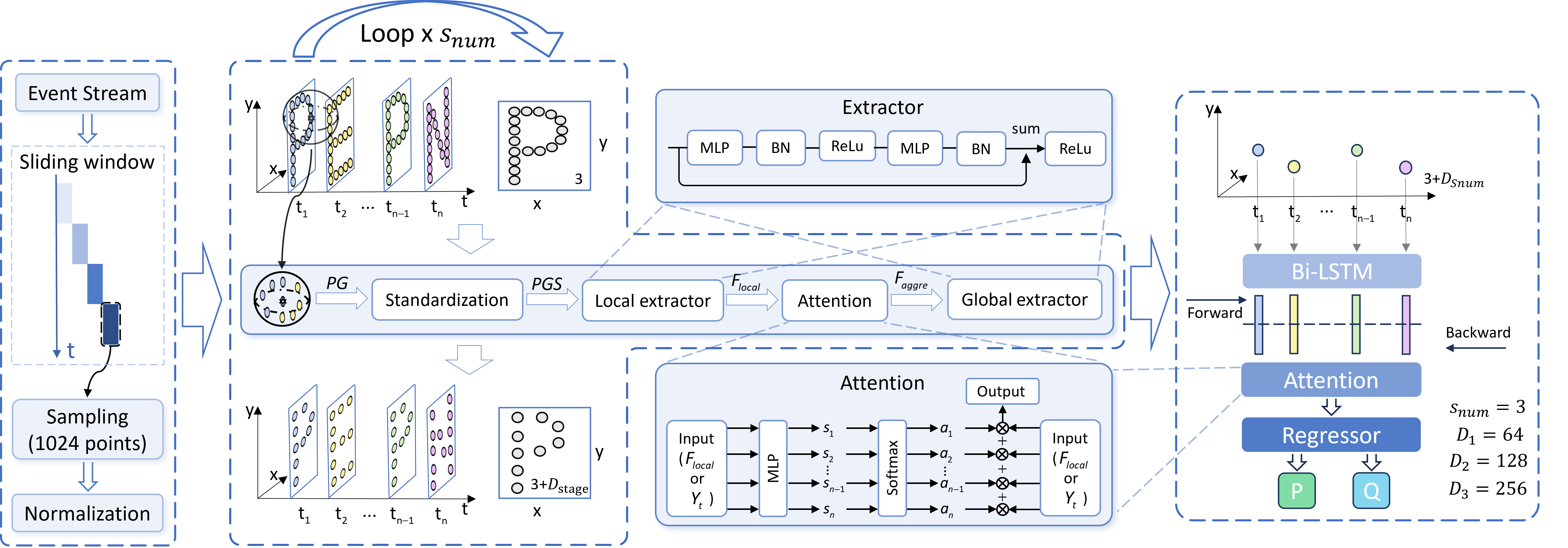}}
\caption{PEPNet overall architecture (the time resolution of $t_1, t_2,... t_n$ is $1\mu s$). The input Event Cloud undergoes direct handling through a sliding window, sampling, and normalization, eliminating the need for any format conversion. Sequentially, the input passes through $S_{num}$ hierarchy structures for spatial feature abstraction and extraction. It further traverses a bidirectional LSTM for temporal feature extraction, culminating in a regressor responsible for 6-DOFs camera pose relocalization.}
\label{fig: pepnet}
\vspace{-0.5cm}
\end{figure*}

PEPNet pipeline consists of four essential modules: (1) a preprocessing module for the original Event Cloud, (2) a hierarchical Point Cloud feature extraction structure, (3) an Attentive Bi-directional LSTM, and (4) a 6-DOFs pose regressor, as illustrated in \cref{fig: pepnet}. In the following sections, we will provide detailed descriptions and formulations for each module.
\subsection{Event Cloud}
To preserve the fine-grained temporal information and original data distribution attributes from the Event Cloud, the 2D-spatial and 1D-temporal event information is constructed into a three-dimensional representation to be processed in Point Cloud. Event Cloud consists of time-series data capturing spatial intensity changes of images in chronological order, and an individual event is denoted as $e_k=(x_k, y_k, t_k, p_k)$, where $k$ is the index representing the $k_{th}$ element in the sequence. Consequently, the set of events within a single sequence ($\mathcal{E}$) in the dataset can be expressed as:
\begin{equation}
        \mathcal{E} = \left\{e_k=(x_k,y_k,t_k,p_k) \mid k=1, \ldots, n\right\}
\end{equation}
For a given pose in the dataset, the ground truth resolution is limited to 5 $ms$, while the event resolution is 1 $\mu s$. Therefore, it is necessary to acquire the events that transpire within the time period we call it sliding window corresponding to the poses, which will serve as the input for the model, as depicted by the following equation:
\begin{equation}
    P_i = \{e_{j \rightarrow l} \mid t_l - t_j = R \} \quad i=1,\ldots,M
\end{equation}
The symbol $R$ represents the time interval of the sliding window, where $j$ and $l$ denote the start and end event index of the sequence, respectively. The variable $M$ represents the number of sliding windows into which the sequence of events $\mathcal{E}$ is divided.
Before being fed into the neural network, $P_i$ also needs to undergo sampling and normalization. Sampling is to unify the number of points $N$ as network inputs. We set $N=1024$ in PEPNet. Additionally, as the spatial coordinates are normalized by the camera’s resolution $w$ and $h$. The normalization process is described by the following equation: 
\begin{equation}
    PN_i = ( \frac{X_i}{w},\frac{Y_i}{h},\frac{T_i-t_j}{t_l - t_j})   
\end{equation}
\begin{equation}
    X_i, Y_i, T_i = \{x_j, \dots, x_l\}, \{y_j, \dots, y_l\}, \{t_j, \dots, t_l\}
\end{equation}
The $X, Y$ is divided by the resolution of the event camera. To normalize $T$, we subtract the smallest timestamp $t_j$ of the window and divide it by the time difference $t_l - t_j$, where $t_l$ represents the largest timestamp within the window. After pre-processing, Event Cloud is converted into the pseudo-Point Cloud, which comprises explicit spatial information $(x,y)$ and implicit temporal information $t$.

\subsection{Hierarchy Structure}
The hierarchy structure is the backbone for processing the pseudo-3D Point Cloud and is composed of four primary modules: grouping and sampling, standardization, feature extractor, and aggregation, as described in the following subsection. To efficiently extract deeper explicit spatial and implicit temporal features, the hierarchical structure is tailored and differs from conventional hierarchical structure in a few ways: First, we no longer force permutation invariance as usually done in mainstream point-based methods \citep{qi2017pointnet,ma2021rethinking}, as the motion information is inherently related to the sequential order of events. Instead, we \textbf{keep the sequence of all events strictly in the same order} as they are generated to preserve the temporal information to be used in the next stage. Second, we replace MaxPooling in aggregation and deploy temporal aggregation which leverages the attention mechanism with softmax, which improves the effective assimilation of temporal information into the resultant feature vectors.

\subsubsection{Grouping and Sampling}
Aligned with the frame-based design concept, our focus is to capture both local and global information. Local information is acquired by leveraging Farthest Point Sampling (FPS) and K-Nearest Neighbors (KNN), while global information is obtained through a dedicated aggregation module.
\begin{equation}
    PS_{i}=FPS(PN_i) \quad
    PG_{i}=KNN(PN_i,PS_{i})
\end{equation}
The input dimension $PN_i$ is $[N,3+D]$, and the centroid dimension $PS_{i}$ is $[N^{'},3+D]$ and the group dimension $PG_{i}$ is $[N^{'}, K,3+2*D]$. $K$ represents the nearest $K$ points of the center point (centroid), D is the feature dimension of the points of the current stage, and 3 is the most original $(X, Y, T)$ coordinate value. Importantly, it should be noted that the ordering of all points in the grouping and sampling process strictly adheres to the timestamp $(T)$, and the dimension $2*D$ of the points in the group is the result of being concatenated to the centroid.
\subsubsection{Standardization}
Next, each group undergoes a standardization process to ensure consistent variability between points within the group, as illustrated in this formula:
\begin{equation}
    PGS_i=\frac{PG_i-PS_i}{Std(PG_i)} \quad Std(PG_i)=\sqrt{\frac{ {\textstyle \sum_{j=0}^{3n-1}} (g_j - \bar g)^2}{3n-1} }
\end{equation}
\begin{equation}
    g =[x_0,y_0,t_0,\dots,x_n,y_n,t_n] 
\end{equation}
Where $PG_i$ and $PS_i$ are the subsets of $PG$ and $PS$, $Std$ is the standard deviation, the dimension of $Std(PG)$ is $M$ which is consistent with the number of sliding windows, and $g$ is the set of coordinates of all points in the $PG_i$.
\subsubsection{Feature extractor}
Following the standardization of $PG$ by dividing the variance by the subtracted mean, the feature extraction is performed using a Multi-Layer Perceptron (MLP) with a residual connection. This process encompasses two steps: local feature extraction and global feature extraction. The feature extractor with a bottleneck can be mathematically represented as:
\begin{align}
     I(x) =& f(\text{BN}(\text{MLP}_1(x))) \\
     O(x) &= \text{BN}(\text{MLP}_2(x)) \\
     Ext(x)& =  f(x + O(I(x)))
\end{align}
BN represents batch normalization layer, while $f$ signifies the nonlinear activation function.
Both local feature extraction and global feature extraction maintain identical input and output dimensions. The dimension increase occurs solely when combining the feature dimension D of the current point with the feature dimension $D$ of the centroid during grouping, resulting in a final dimension of $2*D$. The feature extractor takes an input dimension of $[B, N, K, D]$, and following local feature extraction, the dimension remains $[B, N, K, D]$, $B$ represents batch size. We adopt the attention mechanism for aggregation, yielding an aggregated feature dimension of $[B, N, D]$. Subsequently, the aggregated feature map is then processed through the global feature extractor, completing the feature extraction for the current stage.
\subsubsection{Temporal Aggregation}
Conventional Point Cloud methods favor MaxPooling operations for feature aggregation because it is efficient in extracting the feature from one point among a group of points and discarding the rest. However, MaxPooling involves extracting only the maximum value along each dimension of the temporal axis. It is robust to noise perturbation but also ignores the temporal nuances embedded within the features. Conversely, the integration of attention mechanisms enhances the preservation of those nuanced and useful temporal attributes by aggregating features along the temporal axis through the attention value. To provide a more comprehensive exposition, we employ a direct attention mechanism within the $K$ temporal 
dimensions to effectively aggregate features as shown in \cref{fig: pepnet}. This mechanism enables the explicit integration of temporal attributes, capitalizing on the inherent strict ordering of the $K$ points. The ensuing formula succinctly elucidates the essence of this attention mechanism:
\begin{equation}
    F_{\text{local}} = Ext(x) = (F_{t1},F_{t2},\dots,F_{tk})
\end{equation}
\begin{equation}
    A = \text{SoftMax}(\text{MLP}(F_{\text{local}})) = (a_{t1},a_{t2},\dots,a_{tk})
\end{equation}
\begin{equation}
    F_{\text{aggre}} = A \cdot F_{\text{local}} = F_{t1}\cdot a_{t1} + F_{t2}\cdot a_{t2} + \dots + F_{tk}\cdot a_{tk}
\end{equation}
Upon the application of the local feature extractor, the ensuing features are denoted as $F_{\text{local}}$, and $F_{tk}$ mean the extracted feature of $k_{th}$ point in a group. The attention mechanism comprises an MLP layer with an input layer dimension of $D$ and an output $a_{tk}$ dimension of 1, along with softmax layers. Subsequently, the attention mechanism computes attention values, represented as $A$.  These attention values are then multiplied with the original features through batch matrix multiplication, resulting in the aggregated feature $F_{\text{aggre}}$.

\subsection{A-Bi-LSTM}
The temporal features extracted through the hierarchical structure are independent and parallel, lacking recurrent mechanisms within the network. This distinctive attribute, referred to as 'implicit', contrasts with the conventional treatment of temporal information as an indexed process. Consequently, implicit temporal features \textbf{inadequately capture the interrelations among events along the timeline}, whereas explicit temporal features assume a pivotal role in facilitating the CPR task.  To explicitly capture temporal patterns, we introduce the LSTM network, which has been proven effective in learning temporal dependencies. For optimal network performance, controlled feature dimensionality, and comprehensive capture of bidirectional relationships in pose context, we adopt a bi-directional LSTM network with a lightweight design. The regressor attentively focuses on the output of Bi-LSTM at each timestep and is more inclined towards the start and end features as demonstrated in \cref{figure: attention value}. The integration of bidirectional connections into the recurrent neural network (RNN) is succinctly presented through the following equation:
\begin{align}
    &\mathbf{h}_t = f(\mathbf{W}_h \cdot \mathbf{x}_t + \mathbf{U}_h \cdot \mathbf{h}_{t-1} + \mathbf{b}_h) \\
    &\mathbf{h}'_t = f(\mathbf{W}'_h \cdot \mathbf{x}_t + \mathbf{U}'_h \cdot \mathbf{h}'_{t+1} + \mathbf{b}'_h) \\
    &\mathbf{y}_t =\mathbf{V} \cdot \mathbf{h}_t + \mathbf{b}_y \quad
    \mathbf{y}'_t = \mathbf{V}' \cdot \mathbf{h}'_t + \mathbf{b}'_y
\end{align}
$\mathbf{x}_t$ represents the feature vector at the $t$-th time step of the input sequence, while $\mathbf{h}_{t-1}$ and $\mathbf{h}'_{t+1}$ correspond to the hidden states of the forward and backward RNN units, respectively, from the previous time step. The matrices $\mathbf{W}_h$, $\mathbf{U}_h$, and $\mathbf{b}_h$ denote the weight matrix and bias vector of the forward RNN unit, while $\mathbf{V}$ and $\mathbf{b}_y$ represent the weight matrix and bias vector of its output layer. Similarly, $\mathbf{W}'_h$, $\mathbf{U}'_h$, and $\mathbf{b}'_h$ are associated with the weight matrix and bias vector of the backward RNN unit, and $\mathbf{V}'$ and $\mathbf{b}'_y$ pertain to the weight matrix and bias vector of its output layer. The activation function, denoted as $f(\cdot)$, can be chosen as sigmoid or tanh or other functions. The final output $Y_a$ is aggregated at each moment using the attention mechanism, and $\oplus$ means concat operation.
\begin{align}
    Y_t =& y_t \oplus y'_t \\
    A = \text{SoftMax}&(\text{MLP}(Y_t))  \\
    Y_a =& A \cdot Y_t
\end{align}

\subsection{Loss Function}
 A fully connected layer with a hidden layer is employed to address the final 6-DOFs pose regression task. The displacement vector of the regression is denoted as $\hat{p}$ representing the magnitude and direction of movement, while the rotational Euler angles are denoted as $\hat{q}$ indicating the rotational orientation in three-dimensional space.
\begin{equation}
    \text{Loss} =  \alpha ||\hat p - p ||_2 + \beta ||\hat q -q||_2 
     +\lambda{\textstyle \sum_{i=0}^{n}}  w_i^2
\end{equation}
$p$ and $q$ represent the ground truth obtained from the dataset, while $\alpha$, $\beta$, and $\lambda$ serve as weight proportion coefficients. In order to tackle the prominent concern of overfitting, especially in the end-to-end setting, we incorporate the L2 regularization into the loss function. This regularization, implemented as the second paradigm for the network weights $w$, effectively mitigates overfitting.
\subsection{Overall Architecture}
Next, we will present the PEPNet pipeline in pseudo-code, utilizing the previously defined variables and formulas as described in \cref{algori: 1}.
\begin{algorithm}[t]
\caption{PEPNet pipeline}
\textbf{Input}: Raw event stream $\mathcal{E}$\\
\textbf{Parameters}: $N_{p}=1024, R=1e+3, S_{\text{num}}=3$,$K=24$\\
\textbf{Output}: 6-DOFs pose $(\hat{p},\hat{q})$
\begin{algorithmic}[1] 
\STATE \textbf{Preprocessing} 
\FOR{ $j$ \textbf{in} len($\mathcal{E}$)}
\STATE $P_i.\text{append}(e_{j\rightarrow l})$ ; $j=l$; where $t_l-t_j=R$
\STATE \textbf{if} ($len(P_i)>N_p$): $i=i+1$;
\ENDFOR
\STATE $PN = \text{Normalize}(\text{Sampling}(P))$
\STATE \textbf{Hierarchy structure}
\FOR{ stage in $\text{range}(S_{\text{num}})$}
\STATE \textbf{Grouping and Sampling}($PN$)
\STATE Get $PGS \in [B, N_{\text{stage}}, K, 2*D_{\text{stage}-1}]$
\STATE \textbf{Local Extractor}($PGS$)
\STATE Get $F_{\text{local}} \in [B,N_{\text{stage}},K,D_{\text{stage}}]$
\STATE \textbf{Attentive Aggregate}($F_{\text{local}}$)
\STATE Get $F_{\text{aggre}} \in [B,N_{\text{stage}},D_{\text{stage}}]$
\STATE \textbf{Global Extractor}($F_{\text{aggre}}$)
\STATE Get $PN=F_{\text{global}} \in [B, N_{\text{stage}}, D_{\text{stage}}]$
\ENDFOR
\STATE\textbf{A-Bi-LSTM }
\STATE Forward Get $y_t \in [B, N_{3},D_{S_{\text{num}}}/2]$
\STATE Reverse Get $y'_t \in [B, N_{3},D_{S_{\text{num}}}/2]$
\STATE Attention Get $Y_a \in [B, D_{S_{\text{num}}}]$
\STATE \textbf{Regressor } 
\STATE Get 6-DOFs pose $(\hat{p},\hat{q})$
\end{algorithmic}
\label{algori: 1}
\end{algorithm}
\section{Experiment}
In this section, we present an extensive and in-depth analysis of PEPNet's performance on both indoor and outdoor datasets, encompassing evaluations based on rotational and translational mean squared error (MSE), model parameters, floating-point operations (FLOPs), and inference time. 
PEPNet's training and testing are performed on a server furnished with an AMD Ryzen 7950X CPU, an RTX GeForce 4090 GPU, and 32GB of memory.
\begin{figure*}[t]
\centerline{\includegraphics[width=18cm]{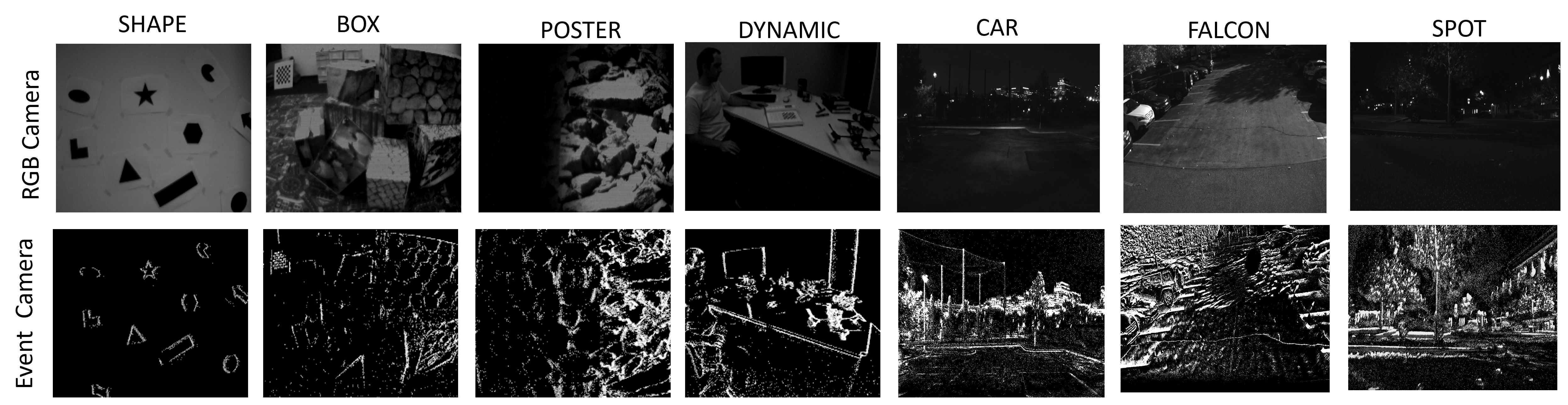}}
\caption{Event-based CPR Dataset visualization.}
\label{fig: visualization of dataset}
\end{figure*}

\subsection{Dataset}
We employ the widely evaluated event-based CPR dataset IJRR \citep{mueggler2017event} and M3ED \citep{chaney2023m3ed}
, encompassing both indoor and outdoor scenes. 
Two distinct methods to partition the CPR dataset \citep{nguyen2019real} have been benchmarked: random split and novel split. In the random split approach, the dataset is randomly selected 70\% of all sequences for training and allocated the remaining sequences for testing. On the other hand, in the novel split, we divide the data chronologically, using the initial 70\% of sequences for training and the subsequent 30\% for testing. 

\subsection{Baseline}
We perform a thorough evaluation of our proposed method by comparing it with SOTA event-based approaches, namely CNN-LSTM \citep{tabia2022deep} and AECRN \citep{lin20226}. Moreover, we present results derived from other well-established computer vision methods, including PoseNet\citep{kendall2015posenet}, Bayesian PoseNet \citep{kendall2016modelling}, Pairwise-CNN \citep{laskar2017camera}, LSTM-Pose \citep{walch2017image}, and SP-LSTM\citep{nguyen2019real}. 

\begin{table*}
\centering
\renewcommand\arraystretch{1.2}
\scalebox{0.75}{
\begin{tabular}{ccccccc|ccllllllllll}
\cline{1-9}
Network             & PoseNet      & Bayesian PoseNet & Pairwise-CNN & LSTM-Pose    & SP-LSTM      & CNN-LSTM     & PEPNet                & $\text{PEPNet}_{\text{tiny}}$   &  &  &  &  &  &  &  &  &  &  \\ \cline{1-9}
Parameter           & 12.43M       & 22.35M           & 22.34M       & 16.05M       & 135.25M      & 12.63M       & \underline{0.774M}                & \textbf{0.064M} &  &  &  &  &  &  &  &  &  &  \\
FLOPs               & 1.584G       & 3.679G           & 7.359G       & 1.822G       & 15.623G      & 1.998G       & \underline{0.459G}                & \textbf{0.033G} &  &  &  &  &  &  &  &  &  &  \\ \cline{1-9}
shapes\_rotation    & 0.109m,7.388$^\circ$ & 0.142m,9.557$^\circ$     & 0.095m,6.332$^\circ$ & 0.032m,4.439$^\circ$ & 0.025m,2.256$^\circ$ & 0.012m,1.652$^\circ$ & \textbf{0.005m,1.372$^\circ$} & \underline{0.006m,1.592$^\circ$}    &  &  &  &  &  &  &  &  &  &  \\
box\_translation    & 0.193m,6.977$^\circ$ & 0.190m,6.636$^\circ$     & 0.178m,6.153$^\circ$ & 0.083m,6.215$^\circ$ & 0.036m,2.195$^\circ$ & \textbf{0.013m},\underline{0.873$^\circ$} & \underline{0.017m},\textbf{0.845$^\circ$} & 0.031m,1.516$^\circ$    &  &  &  &  &  &  &  &  &  &  \\
shapes\_translation & 0.238m,6.001$^\circ$ & 0.264m,6.235$^\circ$     & 0.201m,5.146$^\circ$ & 0.056m,5.018$^\circ$ & 0.035m,2.117$^\circ$ & 0.020m,1.471$^\circ$ & \textbf{0.011m,0.582$^\circ$} & \underline{0.013m, 0.769$^\circ$}   &  &  &  &  &  &  &  &  &  &  \\
dynamic\_6dof       & 0.297m,9.332$^\circ$ & 0.296m,8.963$^\circ$     & 0.245m,5.962$^\circ$ & 0.097m,6.732$^\circ$ & 0.031m,2.047$^\circ$ & \underline{0.016m},1.662$^\circ$ & \textbf{0.015m,1.045$^\circ$} & 0.018m,\underline{1.144$^\circ$}    &  &  &  &  &  &  &  &  &  &  \\
hdr\_poster         & 0.282m,8.513$^\circ$ & 0.290m,8.710$^\circ$     & 0.232m,7.234$^\circ$ & 0.108m,6.186$^\circ$ & 0.051m,3.354$^\circ$ & 0.033m,2.421$^\circ$ & \textbf{0.016m,0.991$^\circ$} & \underline{0.028m,1.863$^\circ$}    &  &  &  &  &  &  &  &  &  &  \\
poster\_translation & 0.266m,6.516$^\circ$ & 0.264m,5.459$^\circ$     & 0.211m,6.439$^\circ$ & 0.079m,5.734$^\circ$ & 0.036m,2.074$^\circ$ & 0.020m,1.468$^\circ$ & \textbf{0.012m,0.588$^\circ$} & \underline{0.019m,0.953$^\circ$}    &  &  &  &  &  &  &  &  &  &  \\ \cline{1-9}
Average             & 0.231m,7.455$^\circ$ & 0.241m,7.593$^\circ$     & 0.194m,6.211$^\circ$ & 0.076m,5.721$^\circ$ & 0.036m,2.341$^\circ$ & 0.019m,1.591$^\circ$ & \textbf{0.013m,0.904$^\circ$} & \underline{0.019m,1.306$^\circ$}    &  &  &  &  &  &  &  &  &  &  \\ \cline{1-9}
\end{tabular}}
\caption{IJRR random split results. The table presents the median error for each sequence, as well as the average error across the six sequences. It also presents the number of parameters and FLOPs for each model.
Bold indicates the most advanced result, while underline signifies the second-best result.}
\label{table:random split}
\end{table*}

\subsection{IJRR Dataset Results}
\subsubsection{Random Split Results}
Based on the findings presented in \cref{table:random split}, it is apparent that PEPNet surpasses other models concerning both rotation and translation errors across all sequences. Notably, PEPNet achieves these impressive results despite utilizing significantly fewer model parameters and FLOPs compared to the frame-based approach. Moreover, PEPNet not only exhibits a remarkable 38\% improvement in the average error compared to the SOTA CNN-LSTM method but also attains superior results across nearly all sequences.
In addressing the more intricate and challenging hdr\_poster sequences, while the frame-based approach relies on a denoising network to yield improved results \citep{jin20216}, PEPNet excels by achieving remarkable performance without any additional processing. This observation strongly implies that PEPNet's Point Cloud approach exhibits greater robustness compared to the frame-based method, highlighting its inherent superiority in handling complex scenarios.

Furthermore, we introduce an alternative variant, PEPNet$_{tiny}$, which integrates a lighter model architecture while preserving relatively strong performance. As depicted in \cref{fig: pepnet}, PEPNet consists of three stages, and the model's size is contingent upon the dimensionality of MLPs at each stage. The dimensions for the standard structure are [64, 128, 256], whereas those for the tiny structure are [16, 32, 64]. As indicated in \cref{table:random split}, even with a mere 0.5\% of the CNN-LSTM's parameter, PEPNet$_{tiny}$ achieves comparable and even slightly superior results. This remarkable outcome emphasizes the superiority of leveraging event cloud data processing directly. 

\begin{table*}
\centering
\renewcommand\arraystretch{1.2}
\scalebox{0.7}{
\begin{tabular}{ccccccccc}
\hline
Network             & PoseNet      & Bayesian PoseNet & Pairwise-CNN  & LSTM-Pose    & SP-LSTM   & DSAC*   & AECRN      & \textbf{PEPNet}        \\ \hline
shapes\_rotation    & 0.201m,12.499$^\circ$ & 0.164m,12.188$^\circ$    & 0.187m,10.426$^\circ$ & 0.061m,7.625$^\circ$ & 0.045m,5.017$^\circ$ &0.029m,2.3$^\circ$ & \underline{0.025m},\underline{2.0$^\circ$} & \textbf{0.016m},\textbf{1.745$^\circ$}  \\
shapes\_translation & 0.198m,6.696$^\circ$  & 0.213m,7.441$^\circ$     & 0.225m,11.627$^\circ$ & 0.108m,8.468$^\circ$ & 0.072m,4.496$^\circ$ &0.038m,2.2$^\circ$ & \underline{0.029m},\underline{1.7$^\circ$} & \textbf{0.026m},\textbf{1.659$^\circ$}  \\
shapes\_6dof        & 0.320m,13.733$^\circ$ & 0.326m,13.296$^\circ$    & 0.314m,13.245$^\circ$ & 0.096m,8.973$^\circ$ & 0.078m,5.524$^\circ$ &0.054m,3.1$^\circ$ & \underline{0.052m},\underline{3.0$^\circ$} & \textbf{0.045m},\textbf{2.984$^\circ$}  \\ \hline
Average             & 0.240m,11.067$^\circ$ & 0.234m,10.975$^\circ$    & 0.242m,11.766$^\circ$ & 0.088m,8.355$^\circ$ & 0.065m,5.012$^\circ$ &0.040m,2.53$^\circ$ & \underline{0.035m},\underline{2.23$^\circ$} & \textbf{0.029m},\textbf{2.13$^\circ$} \\
Inference time      & 5ms           & 6ms              & 12ms          & 9.49ms       & 4.79ms      &30ms  & 30ms      & \textbf{6.7ms}         \\ \hline
\end{tabular}}
\caption{IJRR novel split results. Referred to as \cref{table:random split}, showcases identical information. To assess the model's runtime, we conduct tests on a server platform, specifically focusing on the average time required for inference on a single sample.}
\label{table:novel}
\end{table*}

\begin{figure}
\centering
\includegraphics[width=0.8\linewidth]{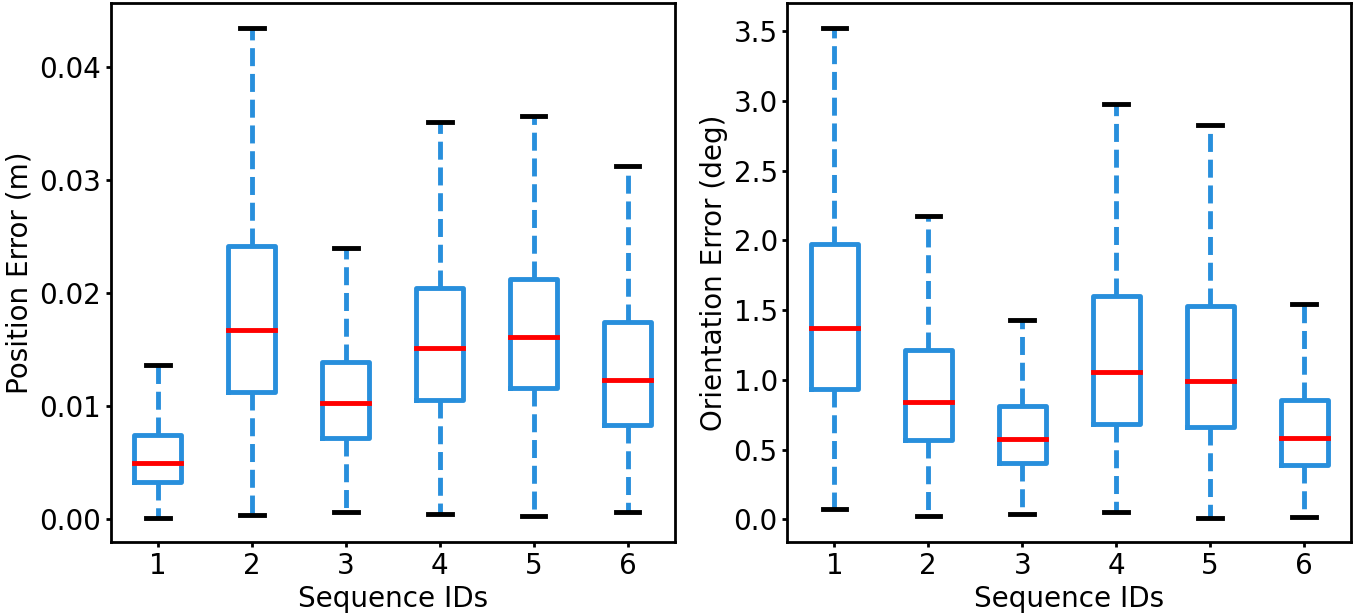}
\caption{Error distribution of event-based CPR results achieved by PEPNet using a random split. (a) Translation errors. (b) Rotation errors.}
\label{fig: error distribution}
\end{figure}
\subsubsection{Error Distribution}
\cref{fig: error distribution} illustrates the error distribution of PEPNet across six distinct sequences using the random split method, specifically: shape rotation, box translation, shape translation, dynamic 6-dof, hdr poster, and poster translation. To enhance clarity, the top and bottom boundaries of the box represent the first and third quartiles, respectively, indicating the inter-quartile range (IQR). The median is denoted by the band within the box. It is observed that the IQR of the translation error approximately locates between 0.004m and 0.024m, while the orientation error ranges from 0.4$^\circ$ to 1.9$^\circ$. 
\subsubsection{Novel Split Results}
To assess the model's robustness, we adopt the novel split as an evaluation criterion, as shown in \cref{table:novel}. During the training process, we observe a more pronounced overfitting phenomenon in PEPNet compared to the random split. We attribute this observation to the disparities in data distributions between the trainset and the testset, as well as the limited data size. Contrary to the methods we compared, PEPNet does not necessitate pre-trained weights. For instance, SP-LSTM relies on pre-trained VGG19 weights from Imagenet, while AECRN requires synthetic heuristic depth and an extensive pretraining process.

To address overfitting, PEPNet employs conventional methods that yield consistent and comparable results with the SOTA on three shape sequences that are displayed in the network column of \cref{table:novel}. It is essential to note that AECRN adopts a hybrid approach, combining neural network regression for scene coordinates with derivable RANSAC for pose estimation. Moreover, this method incurs significant time consumption, with even the SOTA DSAC* algorithm taking nearly 30ms, excluding additional time for data format conversion. This time constraint presents compatibility challenges with the low-latency nature of event cameras. In contrast, PEPNet can execute on a server in just 6.7ms, with the main time-consuming module being grouping and sampling. Furthermore, with potential field programmable gate array (FPGA) or application-specific integrated chip (ASIC) support for these operations\cite{liu2024afpr,fu2024ds}, PEPNet's performance can be further accelerated. 
\begin{table}[h]
\centering
\renewcommand\arraystretch{1.2}
\scalebox{0.6}{
\begin{tabular}{c|ccc|c}
\hline
M3ED                      & PoseNet      & LSTM-Pose    & CNN-LSTM     & \textbf{PEPNet}       \\ \hline
INPUT                     & Event Frame  & Event Frame  & Event frame  & \textbf{Point Cloud}  \\ \hline
Falcon\_Night\_High\_Beams       & 0.181m,2.221$^\circ$ & 0.112m,0.946$^\circ$ & 0.107m,1.435$^\circ$ & \textbf{0.082m,0.575$^\circ$} \\
Car\_Night\_Pen\_S\_Loop         & 1.618m,8.126$^\circ$ & 0.667m,4.914$^\circ$ & 0.773m,3.005$^\circ$ & \textbf{0.577m,1.319$^\circ$} \\
Spot\_Night\_Pen\_Loop           & 1.735m,5.502$^\circ$ & 0.761m,7.898$^\circ$ & \textbf{0.401m},1.771$^\circ$ & \underline{0.468m},\textbf{1.062}$^\circ$ \\
Car\_Pen\_S\_Loop\_darker & 1.841m,4.575$^\circ$ & 0.751m,3.738$^\circ$ & 0.598m,2.772$^\circ$ & \textbf{0.385m,1.01$^\circ$}           \\
Spot\_Plaza\_Light        & 1.372m,9.564$^\circ$ & 0.565m,5.221$^\circ$ & \textbf{0.273}m,2.001$^\circ$ & 
\underline{0.348m},\textbf{1.234}$^\circ$          \\ \hline
Avergae                   & 1.349m,5.998$^\circ$ & 0.571m,4.543$^\circ$ & 0.43m,2.197$^\circ$  & \textbf{0.372m,1.04$^\circ$}  \\ \hline
\end{tabular}
}
\caption{Outdoor extension on M3ED dataset with random split.}
\label{table: outdoor}
\end{table}

\begin{table}
\centering
\renewcommand\arraystretch{1.2}
\scalebox{0.65}{
\begin{tabular}{ccccc|cc|c}
\hline
Condition & HS & LSTM & Bi-LSTM & Aggregation & Translation & Rotation & T+R  \\ \hline
1      & \checkmark                 &      &         &  Max         & 0.015m       & 0.884$^\circ$    & 3.04 \\
2      & \checkmark                    &      &         & Temporal            & 0.014m       & 0.786$^\circ$    & 2.77 \\
3      & \checkmark                    & \checkmark       &         & Max           & 0.014m       & 0.833$^\circ$    & 2.85 \\
4      & \checkmark                    & \checkmark       &         & Temporal           & 0.012m       & 0.603$^\circ$    & 2.25 \\
5      & \checkmark                    &      & \checkmark          & Max          & 0.014m       & 0.813$^\circ$    & 2.82 \\
6      & \checkmark                    &      & \checkmark          & Temporal            & \textbf{0.011m}       & \textbf{0.582$^\circ$}    & \textbf{2.12} \\ \hline
\end{tabular}
}
\caption{Abalation Study for three key modules. T+R = Translation + Rotation$\cdot \pi /180$ (m+rad)}
\label{table: ablation key}
\end{table}
\subsection{M3ED Dataset Results}
We selected three robots (Car, Falcon, and Spot) to extend the application scope of PEPNet across five sequences in an outdoor night setting, as illustrated in the \cref{table: outdoor}. Due to its much higher resolution than IJRR, we performed downsampling processing and more number of points (1024 to 2048), and other experimental configurations are consistent with the IJRR dataset with random split. The results demonstrate the superior performance of PEPNet even in more challenging outdoor environments.
\subsection{Attention Visualization}
\begin{figure}
\centering
\includegraphics[width=0.8\linewidth]{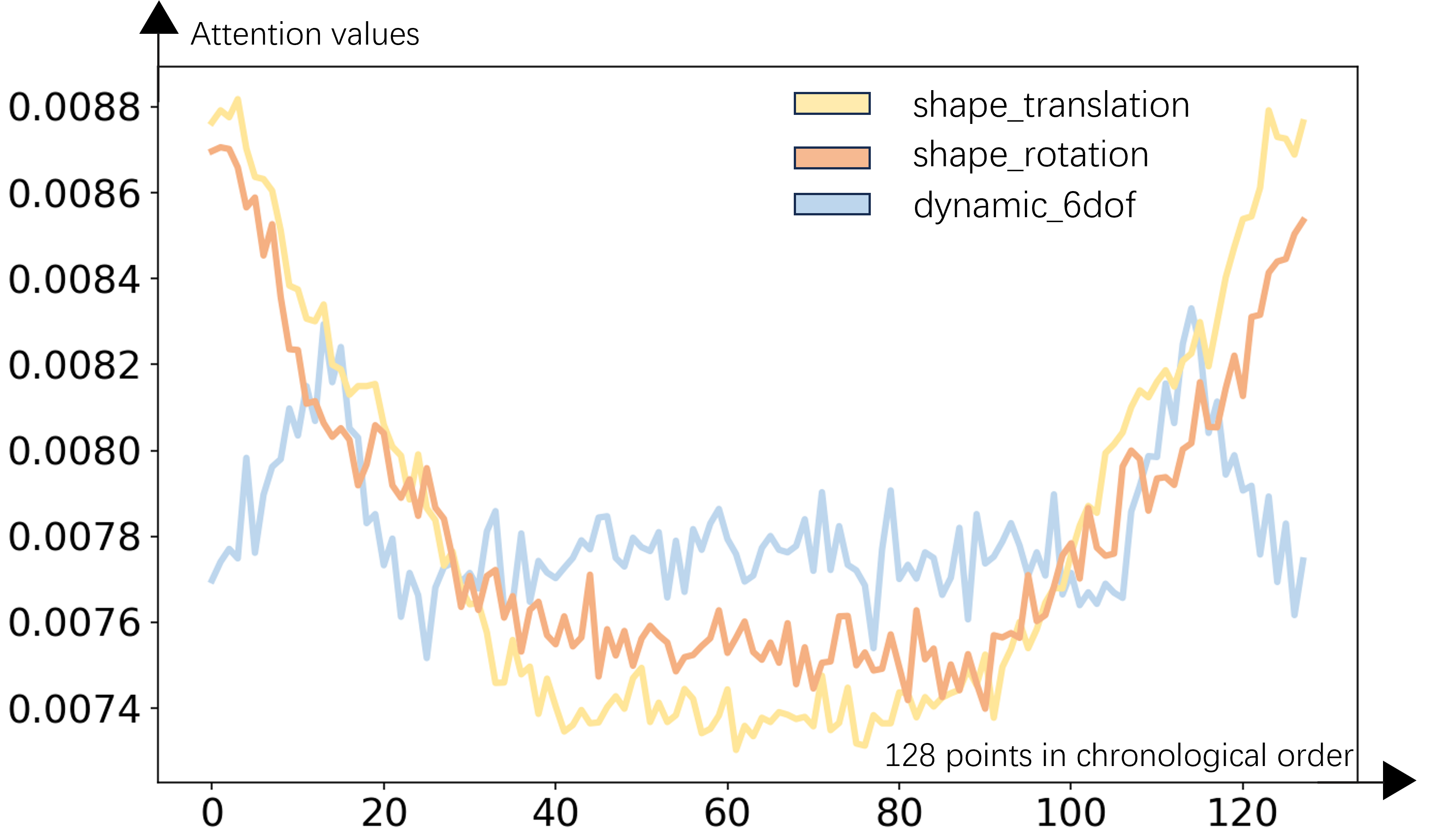}
\caption{Visualization of the attention values in the time domain. 128 points in chronological order on the horizontal axis and the attention values of the corresponding point on the vertical axis.}
\label{figure: attention value}
\end{figure}
As shown in \cref{figure: attention value}, We observe that the attention scores exhibit larger at both the beginning and end. We tentatively infer that the model focuses more on the difference in features between the start and the end for CPR, which is also seen in the geometry approach \cite{mueggler2018continuous,gallego2015event}.

\subsection{Ablation Study}
\textbf{Key Module Ablation:} In order to validate the efficacy of key modules, we conducted an ablation experiment focusing on three primary components: hierarchy structure, Bi-LSTM, and temporal aggregation. 
These experiments are designed to evaluate rotation and translation errors on the shape translation sequence with the random split. 
The combined error (T+R) is measured after processing. Our experimental setup comprises four distinct conditions, as illustrated in \cref{table: ablation key}. Condition 1 represents the sole utilization of the hierarchy structure (HS), while Condition 2 combines the ordinary LSTM. Condition 3 incorporates the bidirectional LSTM, and Condition 4 integrates the attention mechanism for feature aggregation. 
The ablation experiments reveal significant insights. Experiments 1 and 3 demonstrate that augmenting LSTM enhances the extraction of explicit temporal features. Moreover, experiments 3 and 5 reveal the effectiveness of the bidirectional LSTM in extracting motion information. Additionally, experiments 5 and 6 confirm the notable impact of attention in feature aggregation, resulting in a substantial reduction in error rates.

\textbf{Loss ablation:} We incorporated the experiment involving scaling coefficients of the loss function in \cref{table: coefficient in loss}. This experiment utilized a tiny version of PEPNet, trained for 100 epochs, and the outcome is MSE in translation, rotation, and T+R. Across three distinct motion scenarios (translation, rotation, and 6dof) varied coefficient ratios induced deviations in the obtained results. For example, in shape rotation, increasing the weight on rotation makes the results better. 

\begin{table}
\centering
\renewcommand\arraystretch{1.2}
\scalebox{0.6}{
\begin{tabular}{cccc}
\hline
Scence             & $\alpha=0.5,\beta=0.5$               & $\alpha=0.25,\beta=0.75$                        & $\alpha=0.75,\beta=0.25$                         \\ \hline
shape\_translation & \textbf{0.0302m,1.684$^\circ$,5.96} & 0.0359m,1.72$^\circ$,6.59 & 0.0303m,2.056$^\circ$,6.62 \\
shape\_rotation    & 0.0143m,2.888$^\circ$,6.47          & \textbf{0.0159m,2.68$^\circ$,6.27}            & 0.014m,3.36$^\circ$,7.26                       \\
dynamic\_6dof      & \textbf{0.0542m,2.799$^\circ$,10.3} & 0.0611m,2.488$^\circ$,10.5                    & 0.0516m,3.251$^\circ$,10.8                     \\ \hline
\end{tabular}
}
\caption{Abalation Study for loss function's coefficient.}
\label{table: coefficient in loss}
\end{table}
\section{Conclusion}
In this paper, we introduce an end-to-end CPR network that operates directly on raw event clouds without frame-based preprocessing. PEPNet boasts an impressively lightweight framework that adeptly extracts spatial and temporal features, leading to SOTA performance. Diverging from frame-based approaches, our method prioritizes preserving the inherent distribution of the event cloud, capitalizing on its sparse nature to achieve extraordinary capabilities for ultra-low-power applications. 

\textbf{Acknowledgment.} This work was supported in part by the Young Scientists Fund of the National Natural Science Foundation of China (Grant 62305278), as well as the Hong Kong University of Science and Technology (Guangzhou) Joint Funding Program under Grant 2023A03J0154 and 2024A03J0618.
{
    \small
    \bibliographystyle{ieeenat_fullname}
    \bibliography{main}

\begin{thebibliography}{44}
\providecommand{\natexlab}[1]{#1}
\providecommand{\url}[1]{\texttt{#1}}
\expandafter\ifx\csname urlstyle\endcsname\relax
  \providecommand{\doi}[1]{doi: #1}\else
  \providecommand{\doi}{doi: \begingroup \urlstyle{rm}\Url}\fi

\bibitem[Balntas et~al.(2018)Balntas, Li, and Prisacariu]{balntas2018relocnet}
Vassileios Balntas, Shuda Li, and Victor Prisacariu.
\newblock Relocnet: Continuous metric learning relocalisation using neural nets.
\newblock In \emph{Proceedings of the European Conference on Computer Vision (ECCV)}, pages 751--767, 2018.

\bibitem[Brachmann and Rother(2021)]{brachmann2021visual}
Eric Brachmann and Carsten Rother.
\newblock Visual camera re-localization from rgb and rgb-d images using dsac.
\newblock \emph{IEEE transactions on pattern analysis and machine intelligence}, 44\penalty0 (9):\penalty0 5847--5865, 2021.

\bibitem[Brachmann et~al.(2017)Brachmann, Krull, Nowozin, Shotton, Michel, Gumhold, and Rother]{brachmann2017dsac}
Eric Brachmann, Alexander Krull, Sebastian Nowozin, Jamie Shotton, Frank Michel, Stefan Gumhold, and Carsten Rother.
\newblock Dsac-differentiable ransac for camera localization.
\newblock In \emph{Proceedings of the IEEE conference on computer vision and pattern recognition}, pages 6684--6692, 2017.

\bibitem[Chaney et~al.(2023)Chaney, Cladera, Wang, Bisulco, Hsieh, Korpela, Kumar, Taylor, and Daniilidis]{chaney2023m3ed}
Kenneth Chaney, Fernando Cladera, Ziyun Wang, Anthony Bisulco, M~Ani Hsieh, Christopher Korpela, Vijay Kumar, Camillo~J Taylor, and Kostas Daniilidis.
\newblock M3ed: Multi-robot, multi-sensor, multi-environment event dataset.
\newblock In \emph{Proceedings of the IEEE/CVF Conference on Computer Vision and Pattern Recognition}, pages 4015--4022, 2023.

\bibitem[Dosovitskiy et~al.(2020)Dosovitskiy, Beyer, Kolesnikov, Weissenborn, Zhai, Unterthiner, Dehghani, Minderer, Heigold, Gelly, et~al.]{dosovitskiy2020image}
Alexey Dosovitskiy, Lucas Beyer, Alexander Kolesnikov, Dirk Weissenborn, Xiaohua Zhai, Thomas Unterthiner, Mostafa Dehghani, Matthias Minderer, Georg Heigold, Sylvain Gelly, et~al.
\newblock An image is worth 16x16 words: Transformers for image recognition at scale.
\newblock In \emph{International Conference on Learning Representations}, 2020.

\bibitem[Fu et~al.(2024)Fu, Huang, Chen, Fu, Ren, Zhou, Peng, Zong, Pan, and Cheng]{fu2024ds}
Haotian Fu, Yulong Huang, Tingran Chen, Chenyi Fu, Hongwei Ren, Yue Zhou, Shouzhong Peng, Zhirui Zong, Biao Pan, and Bojun Cheng.
\newblock Ds-cim: A 40nm asynchronous dual-spike driven, mram compute-in-memory macro for spiking neural network.
\newblock \emph{IEEE Transactions on Circuits and Systems I: Regular Papers}, 2024.

\bibitem[Gallego et~al.(2015)Gallego, Forster, Mueggler, and Scaramuzza]{gallego2015event}
Guillermo Gallego, Christian Forster, Elias Mueggler, and Davide Scaramuzza.
\newblock Event-based camera pose tracking using a generative event model.
\newblock \emph{arXiv preprint arXiv:1510.01972}, 2015.

\bibitem[Gallego et~al.(2017)Gallego, Lund, Mueggler, Rebecq, Delbruck, and Scaramuzza]{gallego2017event}
Guillermo Gallego, Jon~EA Lund, Elias Mueggler, Henri Rebecq, Tobi Delbruck, and Davide Scaramuzza.
\newblock Event-based, 6-dof camera tracking from photometric depth maps.
\newblock \emph{IEEE transactions on pattern analysis and machine intelligence}, 40\penalty0 (10):\penalty0 2402--2412, 2017.

\bibitem[Gallego et~al.(2020)Gallego, Delbr{\"u}ck, Orchard, Bartolozzi, Taba, Censi, Leutenegger, Davison, Conradt, Daniilidis, et~al.]{gallego2020event}
Guillermo Gallego, Tobi Delbr{\"u}ck, Garrick Orchard, Chiara Bartolozzi, Brian Taba, Andrea Censi, Stefan Leutenegger, Andrew~J Davison, J{\"o}rg Conradt, Kostas Daniilidis, et~al.
\newblock Event-based vision: A survey.
\newblock \emph{IEEE transactions on pattern analysis and machine intelligence}, 44\penalty0 (1):\penalty0 154--180, 2020.

\bibitem[Guo et~al.(2020)Guo, Wang, Hu, Liu, Liu, and Bennamoun]{guo2020deep}
Yulan Guo, Hanyun Wang, Qingyong Hu, Hao Liu, Li Liu, and Mohammed Bennamoun.
\newblock Deep learning for 3d point clouds: A survey.
\newblock \emph{IEEE transactions on pattern analysis and machine intelligence}, 43\penalty0 (12):\penalty0 4338--4364, 2020.

\bibitem[He et~al.(2016)He, Zhang, Ren, and Sun]{he2016deep}
Kaiming He, Xiangyu Zhang, Shaoqing Ren, and Jian Sun.
\newblock Deep residual learning for image recognition.
\newblock In \emph{Proceedings of the IEEE conference on computer vision and pattern recognition}, pages 770--778, 2016.

\bibitem[Jin et~al.(2021)Jin, Yu, Li, and Fei]{jin20216}
Yifan Jin, Lei Yu, Guangqiang Li, and Shumin Fei.
\newblock A 6-dofs event-based camera relocalization system by cnn-lstm and image denoising.
\newblock \emph{Expert Systems with Applications}, 170:\penalty0 114535, 2021.

\bibitem[Kendall and Cipolla(2016)]{kendall2016modelling}
Alex Kendall and Roberto Cipolla.
\newblock Modelling uncertainty in deep learning for camera relocalization.
\newblock In \emph{2016 IEEE international conference on Robotics and Automation (ICRA)}, pages 4762--4769. IEEE, 2016.

\bibitem[Kendall et~al.(2015)Kendall, Grimes, and Cipolla]{kendall2015posenet}
Alex Kendall, Matthew Grimes, and Roberto Cipolla.
\newblock Posenet: A convolutional network for real-time 6-dof camera relocalization.
\newblock In \emph{Proceedings of the IEEE international conference on computer vision}, pages 2938--2946, 2015.

\bibitem[Laskar et~al.(2017)Laskar, Melekhov, Kalia, and Kannala]{laskar2017camera}
Zakaria Laskar, Iaroslav Melekhov, Surya Kalia, and Juho Kannala.
\newblock Camera relocalization by computing pairwise relative poses using convolutional neural network.
\newblock In \emph{Proceedings of the IEEE International Conference on Computer Vision Workshops}, pages 929--938, 2017.

\bibitem[LeCun et~al.(2015)LeCun, Bengio, and Hinton]{lecun2015deep}
Yann LeCun, Yoshua Bengio, and Geoffrey Hinton.
\newblock Deep learning.
\newblock \emph{nature}, 521\penalty0 (7553):\penalty0 436--444, 2015.

\bibitem[Lichtsteiner et~al.(2008)Lichtsteiner, Posch, and Delbruck]{lichtsteiner2008128}
Patrick Lichtsteiner, Christoph Posch, and Tobi Delbruck.
\newblock A 128 $\times$ 128 120 db 15 $\mu$s latency asynchronous temporal contrast vision sensor.
\newblock \emph{IEEE journal of solid-state circuits}, 43\penalty0 (2):\penalty0 566--576, 2008.

\bibitem[Lin et~al.(2022)Lin, Li, Xia, Fei, Yin, and Yang]{lin20226}
Hu Lin, Meng Li, Qianchen Xia, Yifeng Fei, Baocai Yin, and Xin Yang.
\newblock 6-dof pose relocalization for event cameras with entropy frame and attention networks.
\newblock In \emph{The 18th ACM SIGGRAPH International Conference on Virtual-Reality Continuum and its Applications in Industry}, pages 1--8, 2022.

\bibitem[Lin et~al.(2019)Lin, Liu, Huang, Wang, Du, Bai, and Lian]{lin2019deep}
Yimin Lin, Zhaoxiang Liu, Jianfeng Huang, Chaopeng Wang, Guoguang Du, Jinqiang Bai, and Shiguo Lian.
\newblock Deep global-relative networks for end-to-end 6-dof visual localization and odometry.
\newblock In \emph{PRICAI 2019: Trends in Artificial Intelligence: 16th Pacific Rim International Conference on Artificial Intelligence, Cuvu, Yanuca Island, Fiji, August 26--30, 2019, Proceedings, Part II}, pages 454--467. Springer, 2019.

\bibitem[Liu et~al.(2024)Liu, Qian, Wu, Ren, Liu, and Ni]{liu2024afpr}
Haobo Liu, Zhengyang Qian, Wei Wu, Hongwei Ren, Zhiwei Liu, and Leibin Ni.
\newblock Afpr-cim: An analog-domain floating-point rram-based compute-in-memory architecture with dynamic range adaptive fp-adc.
\newblock \emph{arXiv preprint arXiv:2402.13798}, 2024.

\bibitem[Ma et~al.(2021)Ma, Qin, You, Ran, and Fu]{ma2021rethinking}
Xu Ma, Can Qin, Haoxuan You, Haoxi Ran, and Yun Fu.
\newblock Rethinking network design and local geometry in point cloud: A simple residual mlp framework.
\newblock In \emph{International Conference on Learning Representations}, 2021.

\bibitem[Mitrokhin et~al.(2020)Mitrokhin, Hua, Fermuller, and Aloimonos]{mitrokhin2020learning}
Anton Mitrokhin, Zhiyuan Hua, Cornelia Fermuller, and Yiannis Aloimonos.
\newblock Learning visual motion segmentation using event surfaces.
\newblock In \emph{Proceedings of the IEEE/CVF Conference on Computer Vision and Pattern Recognition}, pages 14414--14423, 2020.

\bibitem[Mueggler et~al.(2017)Mueggler, Rebecq, Gallego, Delbruck, and Scaramuzza]{mueggler2017event}
Elias Mueggler, Henri Rebecq, Guillermo Gallego, Tobi Delbruck, and Davide Scaramuzza.
\newblock The event-camera dataset and simulator: Event-based data for pose estimation, visual odometry, and slam.
\newblock \emph{The International Journal of Robotics Research}, 36\penalty0 (2):\penalty0 142--149, 2017.

\bibitem[Mueggler et~al.(2018)Mueggler, Gallego, Rebecq, and Scaramuzza]{mueggler2018continuous}
Elias Mueggler, Guillermo Gallego, Henri Rebecq, and Davide Scaramuzza.
\newblock Continuous-time visual-inertial odometry for event cameras.
\newblock \emph{IEEE Transactions on Robotics}, 34\penalty0 (6):\penalty0 1425--1440, 2018.

\bibitem[Naseer and Burgard(2017)]{naseer2017deep}
Tayyab Naseer and Wolfram Burgard.
\newblock Deep regression for monocular camera-based 6-dof global localization in outdoor environments.
\newblock In \emph{2017 IEEE/RSJ International Conference on Intelligent Robots and Systems (IROS)}, pages 1525--1530. IEEE, 2017.

\bibitem[Nguyen et~al.(2019)Nguyen, Do, Caldwell, and Tsagarakis]{nguyen2019real}
Anh Nguyen, Thanh-Toan Do, Darwin~G Caldwell, and Nikos~G Tsagarakis.
\newblock Real-time 6dof pose relocalization for event cameras with stacked spatial lstm networks.
\newblock In \emph{Proceedings of the IEEE/CVF Conference on Computer Vision and Pattern Recognition Workshops}, pages 0--0, 2019.

\bibitem[Posch et~al.(2010)Posch, Matolin, and Wohlgenannt]{posch2010qvga}
Christoph Posch, Daniel Matolin, and Rainer Wohlgenannt.
\newblock A qvga 143 db dynamic range frame-free pwm image sensor with lossless pixel-level video compression and time-domain cds.
\newblock \emph{IEEE Journal of Solid-State Circuits}, 46\penalty0 (1):\penalty0 259--275, 2010.

\bibitem[Qi et~al.(2017{\natexlab{a}})Qi, Su, Mo, and Guibas]{qi2017pointnet}
Charles~R Qi, Hao Su, Kaichun Mo, and Leonidas~J Guibas.
\newblock Pointnet: Deep learning on point sets for 3d classification and segmentation.
\newblock In \emph{Proceedings of the IEEE conference on computer vision and pattern recognition}, pages 652--660, 2017{\natexlab{a}}.

\bibitem[Qi et~al.(2017{\natexlab{b}})Qi, Yi, Su, and Guibas]{qi2017pointnet++}
Charles~Ruizhongtai Qi, Li Yi, Hao Su, and Leonidas~J Guibas.
\newblock Pointnet++: Deep hierarchical feature learning on point sets in a metric space.
\newblock \emph{Advances in neural information processing systems}, 30, 2017{\natexlab{b}}.

\bibitem[Radwan et~al.(2018)Radwan, Valada, and Burgard]{radwan2018vlocnet++}
Noha Radwan, Abhinav Valada, and Wolfram Burgard.
\newblock Vlocnet++: Deep multitask learning for semantic visual localization and odometry.
\newblock \emph{IEEE Robotics and Automation Letters}, 3\penalty0 (4):\penalty0 4407--4414, 2018.

\bibitem[Rebecq et~al.(2018)Rebecq, Gehrig, and Scaramuzza]{rebecq2018esim}
Henri Rebecq, Daniel Gehrig, and Davide Scaramuzza.
\newblock Esim: an open event camera simulator.
\newblock In \emph{Conference on robot learning}, pages 969--982. PMLR, 2018.

\bibitem[Ren et~al.(2023{\natexlab{a}})Ren, Zhou, Fu, Huang, Xu, and Cheng]{ren2023ttpoint}
Hongwei Ren, Yue Zhou, Haotian Fu, Yulong Huang, Renjing Xu, and Bojun Cheng.
\newblock Ttpoint: A tensorized point cloud network for lightweight action recognition with event cameras.
\newblock \emph{arXiv preprint arXiv:2308.09993}, 2023{\natexlab{a}}.

\bibitem[Ren et~al.(2023{\natexlab{b}})Ren, Zhou, Huang, Fu, Lin, Song, and Cheng]{ren2023spikepoint}
Hongwei Ren, Yue Zhou, Yulong Huang, Haotian Fu, Xiaopeng Lin, Jie Song, and Bojun Cheng.
\newblock Spikepoint: An efficient point-based spiking neural network for event cameras action recognition.
\newblock \emph{arXiv preprint arXiv:2310.07189}, 2023{\natexlab{b}}.

\bibitem[Sekikawa et~al.(2019)Sekikawa, Hara, and Saito]{sekikawa2019eventnet}
Yusuke Sekikawa, Kosuke Hara, and Hideo Saito.
\newblock Eventnet: Asynchronous recursive event processing.
\newblock In \emph{Proceedings of the IEEE/CVF conference on computer vision and pattern recognition}, pages 3887--3896, 2019.

\bibitem[Shavit and Ferens(2019)]{shavit2019introduction}
Yoli Shavit and Ron Ferens.
\newblock Introduction to camera pose estimation with deep learning.
\newblock \emph{arXiv preprint arXiv:1907.05272}, 2019.

\bibitem[Simonyan and Zisserman(2014)]{simonyan2014very}
Karen Simonyan and Andrew Zisserman.
\newblock Very deep convolutional networks for large-scale image recognition.
\newblock \emph{arXiv preprint arXiv:1409.1556}, 2014.

\bibitem[Tabia et~al.(2022)Tabia, Bonardi, and Bouchafa]{tabia2022deep}
Ahmed Tabia, Fabien Bonardi, and Samia Bouchafa.
\newblock Deep learning for pose estimation from event camera.
\newblock In \emph{2022 International Conference on Digital Image Computing: Techniques and Applications (DICTA)}, pages 1--7. IEEE, 2022.

\bibitem[Valada et~al.(2018)Valada, Radwan, and Burgard]{valada2018deep}
Abhinav Valada, Noha Radwan, and Wolfram Burgard.
\newblock Deep auxiliary learning for visual localization and odometry.
\newblock In \emph{2018 IEEE international conference on robotics and automation (ICRA)}, pages 6939--6946. IEEE, 2018.

\bibitem[Walch et~al.(2017)Walch, Hazirbas, Leal-Taixe, Sattler, Hilsenbeck, and Cremers]{walch2017image}
Florian Walch, Caner Hazirbas, Laura Leal-Taixe, Torsten Sattler, Sebastian Hilsenbeck, and Daniel Cremers.
\newblock Image-based localization using lstms for structured feature correlation.
\newblock In \emph{Proceedings of the IEEE International Conference on Computer Vision}, pages 627--637, 2017.

\bibitem[Wang et~al.(2019)Wang, Zhang, Yuan, and Lu]{wang2019space}
Qinyi Wang, Yexin Zhang, Junsong Yuan, and Yilong Lu.
\newblock Space-time event clouds for gesture recognition: From rgb cameras to event cameras.
\newblock In \emph{2019 IEEE Winter Conference on Applications of Computer Vision (WACV)}, pages 1826--1835. IEEE, 2019.

\bibitem[Wu et~al.(2017)Wu, Ma, and Hu]{wu2017delving}
Jian Wu, Liwei Ma, and Xiaolin Hu.
\newblock Delving deeper into convolutional neural networks for camera relocalization.
\newblock In \emph{2017 IEEE International Conference on Robotics and Automation (ICRA)}, pages 5644--5651. IEEE, 2017.

\bibitem[Wu et~al.(2019)Wu, Qi, and Fuxin]{wu2019pointconv}
Wenxuan Wu, Zhongang Qi, and Li Fuxin.
\newblock Pointconv: Deep convolutional networks on 3d point clouds.
\newblock In \emph{Proceedings of the IEEE/CVF Conference on Computer Vision and Pattern Recognition}, pages 9621--9630, 2019.

\bibitem[Yang et~al.(2019)Yang, Zhang, Ni, Li, Liu, Zhou, and Tian]{yang2019modeling}
Jiancheng Yang, Qiang Zhang, Bingbing Ni, Linguo Li, Jinxian Liu, Mengdie Zhou, and Qi Tian.
\newblock Modeling point clouds with self-attention and gumbel subset sampling.
\newblock In \emph{Proceedings of the IEEE/CVF conference on computer vision and pattern recognition}, pages 3323--3332, 2019.

\bibitem[Zhao et~al.(2021)Zhao, Jiang, Jia, Torr, and Koltun]{zhao2021point}
Hengshuang Zhao, Li Jiang, Jiaya Jia, Philip~HS Torr, and Vladlen Koltun.
\newblock Point transformer.
\newblock In \emph{Proceedings of the IEEE/CVF International Conference on Computer Vision}, pages 16259--16268, 2021.

\end{thebibliography}
}


\end{document}